\documentclass[aip,preprint]{revtex4-1}
\usepackage{graphicx}
\usepackage[section]{placeins}
 \usepackage{float}
 \usepackage{amsmath}

\begin{document}



\title{Robust and Efficient Interference Neural Networks for Defending Against Adversarial Attacks in ImageNet}

\author{Yunuo Xiong}
\email{xiongyunuo@hbpu.edu.cn}
\affiliation{Center for Fundamental Physics and School of Mathematics and Physics, Hubei Polytechnic University, Huangshi 435003, China}

\author{Shujuan Liu}
\affiliation{Center for Fundamental Physics and School of Mathematics and Physics, Hubei Polytechnic University, Huangshi 435003, China}

\author{Hongwei Xiong}
\email{xionghongwei@hbpu.edu.cn}

\affiliation{Center for Fundamental Physics and School of Mathematics and Physics, Hubei Polytechnic University, Huangshi 435003, China}

\begin{abstract}
The existence of adversarial images has seriously affected the task of image recognition and practical application of deep learning, it is also a key scientific problem that deep learning urgently needs to solve. By far the most effective approach is to train the neural network with a large number of adversarial examples. However, this adversarial training method requires a huge amount of computing resources when applied to ImageNet, and has not yet achieved satisfactory results for high-intensity adversarial attacks. In this paper, we construct an interference neural network by applying additional background images and corresponding labels, and use pre-trained ResNet-152 to efficiently complete the training. Compared with the state-of-the-art results under the PGD attack, it has a better defense effect with much smaller computing resources. This work provides new ideas for academic research and practical applications of effective defense against adversarial attacks.

\end{abstract}

\maketitle

\section{Introduction}

Despite the significant value and practical application of deep learning in image recognition \cite{LeCun}, the key scientific and technical obstacle of adversarial attacks \cite{Sze,Good,Ngu,Sharif,Kurakin,Heaven,Su} has long been a challenge of deep learning. Experiments on ImageNet have shown that even after applying imperceptible specially crafted perturbations, ordinary neural networks will misidentify the adversarial image completely  \cite{Sze}. Since the discovery of adversarial attacks, a large number of different methods have been tried, but no satisfactory defense results have been achieved. 

To date, the most effective method appears to be adversarial training \cite{Madry,Schott,Sha,Bai}, which involves adding a large number of adversarial images to the training of neural networks in the hope that since the neural network has already been exposed to adversarial images, it can now be immune to attacks when recognizing new images with adversarial perturbations. Only when the defenses of ImageNet \cite{deng} are effective, can we resist potential adversarial attacks in practical applications. 
However, the method of adversarial training faces two difficulties for adversarial training for ImageNet \cite{Tramer,Qin,Kannan,Xie2,Balaji,Hendrycks,Wong,Andriu,Lee,HuK}: (1) the adversarial images make training more difficult; (2) even after completing the computationally expensive training, it still does not achieve satisfactory results for strong adversarial attacks.  

This work will demonstrate how to overcome these two difficulties based on interference neural networks through experiments on ImageNet. In interference neural networks, only a few simple background images and corresponding additional labels are added to the images during training. Based on the pre-trained ResNet-152 model, we can complete the training of interference neural networks on a single GPU in 30 hours. To the best of our knowledge, our defense achieves the best results so far under white-box PGD-500 attacks with attack strengths less than 16/255. Experiments also show that even at an attack strength of 96/255, the defense effect at the Top-1 recognition rate can reach 50$\%$.

\section{The background and related work}

We define the loss function of the neural network as $J(\theta,x_j,M_x)$. Here, $\theta$ represents the weight parameters, $x_j$ represents the image, and $M_x$ represents the correct label for $x_j$. The goal of training a neural network is to minimize $J$ so that it can predict the correct label for a given image as accurately as possible. We can now add a small perturbation $\delta x_j$ to $x_j$. The change in the loss function can then be approximated as $\delta J=\sum_j\frac{\partial J}{\partial x_j}\delta x_j$. If we design a special $\delta x_j$ based on the calculation of the gradient of $J$, $\delta J$ can become very large, resulting in misclassification. This is known as an adversarial attack \cite{Sze,Good}. The number of terms in the sum of $\delta J$ is different for different images. Obviously, the more terms, the greater the change in $J$ for a given perturbation of the same magnitude. For MNIST, the number of terms is 784; for CIFAR-10, it is 3072; and for ImageNet, when the input image is $256\times 256\times 3$, it is as high as 196608.
Therefore, even if we find a method to defend against adversarial attacks on MNIST and CIFAR-10, it is not clear whether it will be effective on ImageNet. In practical terms, it is essential to develop methods that can effectively defend against adversarial attacks on ImageNet.

The first relatively effective defense against adversarial attacks was the adversarial training proposed by Madry et al. \cite{Madry} for MNIST. However, for CIFAR-10, the number of terms in the sum of $\delta J$ is four times that of MNIST. Therefore, the defense effect of Madry et al. on CIFAR-10 is much worse than that on MNIST. Another reason for the worse performance on CIFAR-10 is that adversarial training typically requires adding an order of magnitude more adversarial examples to the training set, which makes training more difficult and requires much more computational resources. This is one of the reasons why Madry et al. \cite{Madry} did not study ImageNet.

In any case, with the rapid development of GPU technology, researchers have begun to use adversarial training to address the defense difficulty of ImageNet after the work of Madry et al \cite{Madry}. A representative work is by Balaji et al. \cite{Balaji}, who used 128 Nvidia V100 GPUs to complete adversarial training. Under the PGD-1000 attack, they achieved a Top-1 recognition rate of 40.40$\%$ at an attack intensity of 16/255, and a recognition rate of 57.26$\%$ on clean images. Balaji et al. \cite{Balaji} were not satisfied with this result, and proposed instance adaptive adversarial training (IAAT), which improved the defense effect of adversarial images to a certain extent. In particular, the recognition rate on clean images reached 67.44$\%$. However, compared to traditional adversarial training, the recognition rate decreases faster with the increase of attack intensity. At an attack intensity of 16/255, the test accuracy is only 27.85$\%$, which is much worse than the traditional adversarial training effect.

Although Balaji et al.'s IAAT \cite{Balaji} provides the state-of-the-art defense results for ImageNet, we still need to find new ways to overcome the challenges of huge amount of required computational resources and the sharp decline of recognition rate under high-intensity attacks. Based on the results of many existing attempts \cite{Tramer,Qin,Kannan,Xie2,Balaji,Hendrycks,Wong,Andriu,Lee,HuK} along the line of adversarial training for ImageNet, we have to explore new approaches to address this challenge. Some researchers \cite{quan} have even proposed using quantum computing to solve this problem. Our following research shows that deep learning has the potential to overcome adversarial attacks without the need for quantum computing.

\section{Interference Neural Networks and Their Experimental Results on ImageNet}

It is obvious that the associative law of addition $(x_j+\delta x_j)+y_j=x_j+(\delta x_j+y_j)$ also applies to the input end of a neural network. Interestingly, we found that this simplest and most basic law of mathematics can provide important insights into overcoming adversarial attacks. For the sake of simplicity, we will ignore the subscripts of $x_j$ in the following.

\begin{figure}[htbp]
\begin{center}
 \includegraphics[width=0.65\textwidth]{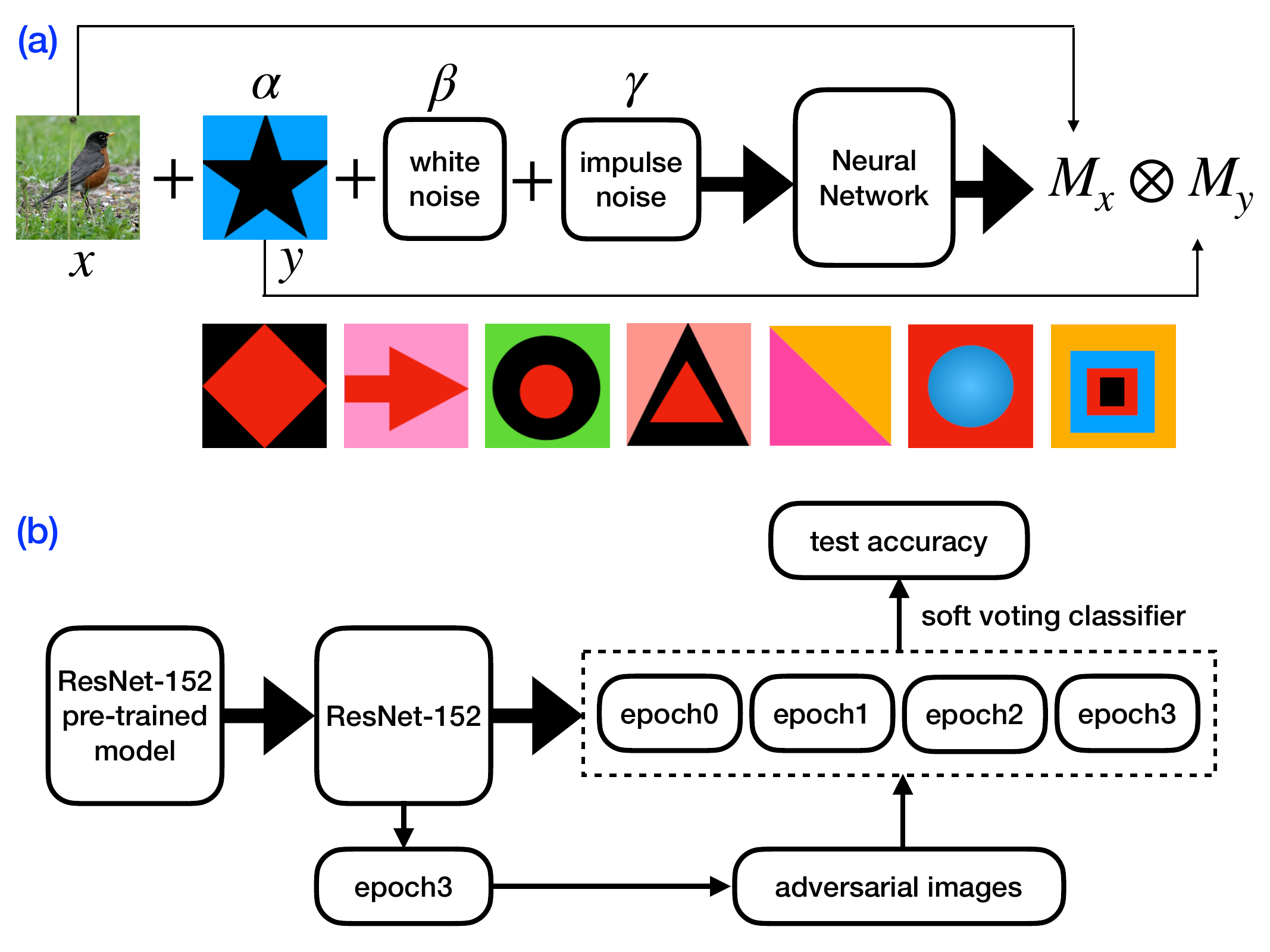} 
\caption{\label{fig1} Figure (a) represents the basic architecture of the interference neural network. Here $x$ represents ImageNet, $y$ represents the background image. $M_x$ and $M_y$ represent the labels of ImageNet and the background image, respectively. Figure (a) shows the 8 artificially made background images used in this work. Figure (b) shows the process of training and identifying adversarial images of the interference neural network. }
\end{center}
\end{figure}

We assume that $x$ represents an ImageNet image, and $y$ represents the background image shown in Figure  \ref{fig1}(a), with their corresponding labels being $M_x$ and $M_y$, respectively. Let us now consider applying some adversarial attack $\delta x$ to $x$. In a normal neural network, we assume that this attack will completely lead to misclassification. However, due to the associative law of addition, when the background image $y$ exists, $\delta x$ can also be combined with $y$. Since ImageNet images and background images are completely different, the combination of $\delta x$ and $y$ does not produce an effective adversarial attack on the background image. The neural network naturally cannot distinguish between the two visually different combinations. Therefore, intuitively, the associative law of addition at the input end leads to the protective effect of the background image on $x$, which is equivalent to the background image $y$ absorbing the adversarial attack $\delta x$. We also found that this protective effect can be further enhanced by applying white noise and salt-and-pepper noise (also known as impulse noise) during training and testing. The reason for this further enhanced protection is that due to the associative law of addition, $\delta x$ can be further absorbed by white noise and salt-and-pepper noise, thus playing a protective role. Experiments show that the joint absorption of $\delta x$ by background images, white noise, and salt-and-pepper noise can achieve excellent defense results.

In Figure \ref{fig1}(a), we give the details of the interference neural network. For any image in the training set, we apply white noise with intensity $\beta$, so that the input becomes $(x+\beta n)/(1+\beta)$. Here $n$ is a random number in $(0,1)$. Then we give the input a salt-and-pepper noise treatment with intensity $\gamma$ to make the input $x'$. Finally, we randomly add one of the background images to make the input $x'+\alpha y$. Here $\alpha$ is the scaling factor. After this processing is applied to each image in the training set, the number of training set images does not change, and for ImageNet it is still 1,281,167. As a result, we have added four additional hyperparameters $(\alpha,\beta,\gamma,K)$ when training the interference neural network. Here $K$ represents the number of background images. In a sense, the background image itself is also a hyperparameter.

In this work, we adopted the training method given in Figure  \ref{fig1}(b). We used the weight parameters of ResNet-152 on ImageNet as a pre-trained model, and then trained it based on the same ResNet-152. Compared with the usual ResNet-152, the number of labels is modified to $1000\times K$ after considering the background image. Experiments show that the pre-trained parameters can be well transferred to the new interference neural network. In Figure  \ref{fig1}(b), we give the basic process of training and attacking. We train for 4 epochs at a time to obtain 4 sets of weight parameters. Using one of the epochs, for example the weight parameters obtained at epoch 3, we calculate the gradient of the loss function on each image in the test set based on the interference neural network to generate PGD-500 adversarial attacks, thus forming adversarial images. During testing, we add the background images, white noise, and salt-and-pepper noise with the same hyperparameters as during training, and then use all four sets of weight parameters to add a soft voting classifier.

\begin{figure}[htbp]
\begin{center}
 \includegraphics[width=0.65\textwidth]{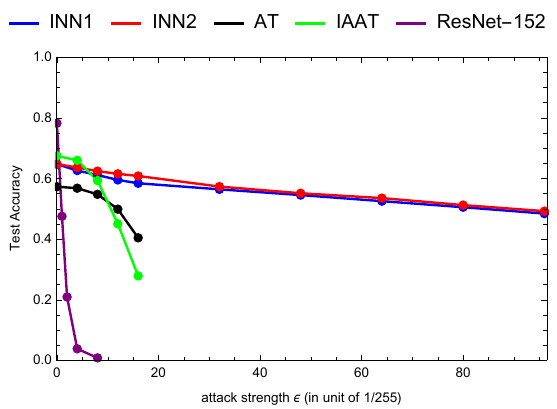} 
\caption{\label{fig2} The blue and red points in the figure represent the defense effects of the interference neural network under two white-box attacks. The black point represents the recognition results of Balaji et al. using adversarial training, and the green point represents the defense results of their improved IAAT \cite{Balaji} under the PGD-1000 attack. The purple point in the figure represents the behavior of ResNet-152 without considering any defense under the PGD-500 attack. }
\end{center}
\end{figure}

In white-box attacks, we need to pay attention to a subtle point: there are several different ways to calculate the gradient of the loss function to generate PGD adversarial attacks $\delta x$. One method is to calculate the gradient based on one of the epochs' weight parameters after stacking the background and applying the same intensity of white noise and salt-and-pepper noise as in training for the test image. Another method is to simply calculate the gradient with the test image and the weight parameters. Experimental results show that there is no significant difference between the two methods, and the first method produces slightly stronger attacks. In the results of Figure  \ref{fig2}, we use hyperparameters of $(\alpha=0.5,\beta=0.4,\gamma=0.4,K=8)$. The specific 8 background images are shown in Figure  \ref{fig1}(a). We use a batch size of 512, momentum=0.9, and weight decay=$5\times 10^{-4}$ when training. In the interference neural network, we use the weight parameters from epoch 3 to prepare adversarial attacks.

In Figure  \ref{fig2}, we show the defense results of the Top-1 recognition rate on the test set under the PGD-500 attack. The horizontal axis in Figure  \ref{fig2} represents different attack strengths $\epsilon/255$, and the vertical axis represents the test accuracy. We used 500 iterations to generate the PGD attack. When calculating $\delta x$, the step size is 2.5 times the $\epsilon/255$ divided by the number of iterations as usual \cite{Madry}. The blue point (INN1) in Figure  \ref{fig2} represents the recognition results of our adversarial attacks on images in the test set, and the background image, white noise, and salt-and-pepper noise are all added to the test image when calculating the gradient in the adversarial attack. The red point (INN2) is the classification accuracy when manufacturing adversarial images based on the test images only. The black point and the green point in Figure  \ref{fig2} represent the recognition results of Balaji et al. \cite{Balaji} based on the usual adversarial training and the improved IAAT, respectively. Since the results of the two methods provided by Balaji et al. both have a sharp decline under the attack strength of $\epsilon>16/255$, they did not provide more data. In our results, the recognition rate only slowly decreases until the high-intensity attack of $\epsilon=96/255$. On average, the recognition rate is also significantly higher than the results of the two methods provided by Balaji et al. \cite{Balaji} in the $(0<\epsilon<16/255)$ interval. As a reference, the purple point in the figure represents the behavior of ResNet-152 without considering any defense under the PGD-500 attack. We can clearly notice the devastating consequences of adversarial attacks on image recognition without defense.

\begin{figure}[htbp]
\begin{center}
 \includegraphics[width=0.65\textwidth]{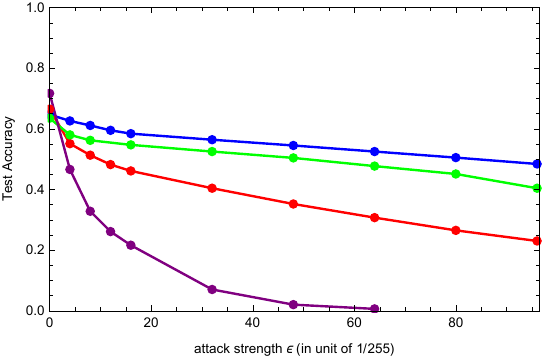} 
\caption{\label{fig3} The figure shows the defense performance of the interference neural network under different hyperparameters $(\alpha,\beta,\gamma,K)$.}
\end{center}
\end{figure}

In Figure \ref{fig3}, we show the defense performance of different hyperparameters. We observe that as the hyperparameters $\alpha,\beta,\gamma$ increase, the robustness to adversarial attacks improves, but the recognition rate of clean images decreases. The hyperparameters of the blue point are $\alpha=0.5,\beta=0.4,\gamma=0.4,K=8$, the green point is $\alpha=0.3,\beta=0.3,\gamma=0.4,K=4$, the red point is $\alpha=0.2,\beta=0.3,\gamma=0.2,K=4$, and the purple point is $\alpha=0.1,\beta=0.1, \gamma=0.1, K=8$. In the preparation of adversarial attacks, we calculate the gradients based on one of the epoch's weight parameters for test images after superimposing the background and applying white noise and salt and pepper noise of the same strength as in the training. In addition, we usually calculate four epochs with the help of a pre-trained model, and use one epoch to generate adversarial examples, and then use four epochs to perform a soft voting classifier.

It is worth noting that the interference neural network does not increase the number of ImageNet training images, so the computational resources required by the interference neural network are much less than those required by adversarial training methods. In the work of Balaji et al., 128 Nvidia V100 GPUs were used for training, and the training took 94 hours \cite{Balaji}. We used the Apple M2 Ultra chip (Mac Studio) with 192G unified memory, and the training time for 4 epochs was 30 hours.

Compared with adversarial training, the reason for high efficiency and high defense performance is that the interference neural network has a good absorption effect on the background image, white noise, and salt-and-pepper noise combined when defending against adversarial attacks. We might be able to understand the effect of the interference neural network through a figurative analogy. A face can be made unrecognizable by high-end makeup. However, a sponge filled with water can absorb the makeup and reveal the true appearance. In the interference neural network, the background image is similar to a sponge, and the white noise and salt-and-pepper noise are similar to water and the sponge's pores. The training and recognition of the interference neural network allow the "sponge" to absorb this makeup.

\section{Conclusion}

In summary, we have applied the interference neural network (INN) to the training and defense of ImageNet. From the defense effect point of view, it is better than the current state-of-the-art results based on adversarial training \cite{Tramer,Qin,Kannan,Xie2,Balaji,Hendrycks,Wong,Andriu,Lee,HuK}, and the use of pre-trained models also reduces the computational resources required by adversarial training by more than one order of magnitude. The successful experiments of INN on ImageNet have the hope of laying the foundation for the practical application of adversarial defense.

Like adversarial training, the decrease in the recognition rate of clean images is a challenge that INN is currently facing. Although for clean images, INN has basically reached the level of IAAT, and has a significant improvement over the usual adversarial training; however, compared with ResNet-152 without considering adversarial defense, there is still more than 10$\%$ of recognition rate reduction in the clean image aspect. 
The recognition rate of the test set of clean images in this article is a few percentage points higher than the accuracy of the famous AlexNet \cite{Krizhevsky} of  Krizhevsky et al. Nevertheless, the decrease in the recognition rate of clean images is a problem that INN needs to overcome in the future. Finally, it is also a topic worth studying in the future whether INN can play a role in natural language processing under adversarial attacks \cite{ZouA}, etc.

\textbf{Acknowledgments}  The authors would like to acknowledge support from Hubei Polytechnic University. The authors also acknowledge the great encouragement and support by Prof. Yu Hongsheng and Prof. Chen Xiangsong.


\textbf{Funding} This work is partly supported by the National Natural Science Foundation of China under grant numbers 11175246, and 11334001. 



\textbf{Availability of data and material}
The data that support the findings of this study are available from the corresponding author upon reasonable request.








\begin{thebibliography}{10}


\bibitem{LeCun} LeCun, Y., Bengio, Y., Hinton, G. (2015). Deep learning. Nature, 521(7553), 436-444.


\bibitem{Sze} Szegedy, C., Zaremba, W., Sutskever, I., Bruna, J., Erhan, D., Goodfellow, I., Fergus, R. (2013). Intriguing properties of neural networks. arXiv preprint arXiv:1312.6199.

\bibitem{Good} Goodfellow, I. J., Shlens, J.,  Szegedy, C. (2014). Explaining and harnessing adversarial examples. arXiv preprint arXiv:1412.6572.

\bibitem{Ngu} Nguyen, A., Yosinski, J., Clune, J. (2015). Deep neural networks are easily fooled: High confidence predictions for unrecognizable images. In Proceedings of the IEEE conference on computer vision and pattern recognition (pp. 427-436).

\bibitem{Sharif} Sharif, M., Bhagavatula, S., Bauer, L.,  Reiter, M. K. (2016). Accessorize to a crime: Real and stealthy attacks on state-of-the-art face recognition. In Proceedings of the 2016 acm sigsac conference on computer and communications security (pp. 1528-1540).

\bibitem{Kurakin} Kurakin, A., Goodfellow, I. J., Bengio, S. (2018). Adversarial examples in the physical world. In Artificial intelligence safety and security (pp. 99-112). Chapman and Hall/CRC.

\bibitem{Heaven} Heaven, D. (2019). Deep trouble for deep learning. Nature, 574 (7777), 163-166.

\bibitem{Su} Su, J., Vargas, D. V., Sakurai, K. (2019). One pixel attack for fooling deep neural networks. IEEE Transactions on Evolutionary Computation, 23(5), 828-841.

\bibitem{Madry} Madry, A., Makelov, A., Schmidt, L., Tsipras, D., Vladu, A. (2017). Towards deep learning models resistant to adversarial attacks. arXiv preprint arXiv:1706.06083.

\bibitem{Schott} Schott, L., Rauber, J., Bethge, M., Brendel, W. (2018). Towards the first adversarially robust neural network model on MNIST. arXiv preprint arXiv:1805.09190.

\bibitem{Sha} Shafahi, A., Najibi, M., Ghiasi, M. A., Xu, Z., Dickerson, J., Studer, C., ...,  Goldstein, T. (2019). Adversarial training for free!. Advances in Neural Information Processing Systems, 32.

\bibitem{Bai} Bai, T., Luo, J., Zhao, J., Wen, B., Wang, Q. (2021). Recent advances in adversarial training for adversarial robustness. arXiv preprint arXiv:2102.01356.


\bibitem{deng} Deng, J., Dong, W., Socher, R., Li, L. J., Li, K.,  Fei-Fei, L. (2009). Imagenet: A large-scale hierarchical image database. In 2009 IEEE conference on computer vision and pattern recognition (pp. 248-255). IEEE.


\bibitem{Tramer} Tramèr, F., Kurakin, A., Papernot, N., Goodfellow, I., Boneh, D., McDaniel, P. (2017). Ensemble adversarial training: Attacks and defenses. arXiv preprint arXiv:1705.07204.

\bibitem{Qin} Qin, C., Martens, J., Gowal, S., Krishnan, D., Dvijotham, K., Fawzi, A., ... Kohli, P. (2019). Adversarial robustness through local linearization. Advances in Neural Information Processing Systems, 32. 

\bibitem{Kannan} Kannan, H., Kurakin, A., Goodfellow, I. (2018). Adversarial logit pairing. arXiv preprint arXiv:1803.06373. 

\bibitem{Xie2} Xie, C., Wu, Y., Maaten, L. V. D., Yuille, A. L., He, K. (2019). Feature denoising for improving adversarial robustness. In Proceedings of the IEEE/CVF conference on computer vision and pattern recognition (pp. 501-509).

\bibitem{Balaji} Balaji, Y., Goldstein, T., Hoffman, J. (2019). Instance adaptive adversarial training: Improved accuracy tradeoffs in neural nets. arXiv preprint arXiv:1910.08051. 

\bibitem{Hendrycks} Hendrycks, D., Mazeika, M., Kadavath, S., Song, D. (2019). Using self-supervised learning can improve model robustness and uncertainty. Advances in neural information processing systems, 32. 

\bibitem{Wong} Wong, E., Rice, L.,  Kolter, J. Z. (2020). Fast is better than free: Revisiting adversarial training. arXiv preprint arXiv:2001.03994. 

\bibitem{Andriu} Andriushchenko, M., Flammarion, N. (2020). Understanding and improving fast adversarial training. Advances in Neural Information Processing Systems, 33, 16048-16059.

\bibitem{Lee} Lee, S., Lee, H., Yoon, S. (2020). Adversarial vertex mixup: Toward better adversarially robust generalization. In Proceedings of the IEEE/CVF Conference on Computer Vision and Pattern Recognition (pp. 272-281).

\bibitem{HuK} Hu, K., Zou, A., Wang, Z., Leino, K., Fredrikson, M. (2023). Scaling in Depth: Unlocking Robustness Certification on ImageNet. arXiv preprint arXiv:2301.12549.


\bibitem{quan} West, M. T., Tsang, S. L., Low, J. S., Hill, C. D., Leckie, C., Hollenberg, L. C., ... Usman, M. (2023). Towards quantum enhanced adversarial robustness in machine learning. Nature Machine Intelligence, 1-9.


\bibitem{Krizhevsky} Krizhevsky, A., Sutskever, I.,  Hinton, G. E. (2012). Imagenet classification with deep convolutional neural networks. Advances in neural information processing systems, 25.

\bibitem{ZouA} Zou, A., Wang, Z., Kolter, J. Z., Fredrikson, M. (2023). Universal and transferable adversarial attacks on aligned language models. arXiv preprint arXiv:2307.15043.

















\end{thebibliography}
\end{document}